\pdfoutput=1

\documentclass[11pt]{article}

\usepackage[final]{acl}    

\usepackage{times}
\usepackage{latexsym}
\usepackage{multirow}
\usepackage{amsthm,amsmath,amssymb}
\usepackage{graphicx}
\usepackage{tabularx}
\usepackage{mathrsfs}
\usepackage{subfigure}
\usepackage[T1]{fontenc}

\usepackage[utf8]{inputenc}

\usepackage{microtype}

\usepackage{inconsolata}

\title{Unsupervised Distractor Generation via Large Language Model Distilling and Counterfactual Contrastive Decoding}

\author{
Fanyi Qu \quad
Hao Sun \quad
Yunfang Wu\thanks{\ \ \ Corresponding author.} \\
  National Key Laboratory for Multimedia Information Processing, Peking University \\
  MOE Key Laboratory of Computational Linguistics, Peking University\\
  School of Computer Science, Peking University \\
  \texttt{\{fanyiqu,2301213218,wuyf\}@pku.edu.cn}
}

\begin{document}
\maketitle
\begin{abstract}
Within the context of reading comprehension, the task of Distractor Generation (DG) aims to generate several incorrect options to confuse readers. 
Traditional supervised methods for DG rely heavily on expensive human-annotated distractor labels. In this paper, we propose an unsupervised DG framework,  
leveraging Large Language Models (LLMs) as cost-effective annotators to enhance the DG capability of smaller student models. Specially, to perform knowledge distilling, we propose a dual task training strategy 
that integrates pseudo distractors from LLMs and 
the original answer information as the 
objective targets with a two-stage training process.
Moreover, 
we devise a counterfactual contrastive decoding mechanism 
for increasing the distracting capability of the DG model. 
Experiments show that our unsupervised generation method with Bart-base greatly surpasses 
GPT-3.5-turbo 
performance with 
only 200$\times$ fewer model parameters. 
Our proposed unsupervised 
DG method 
offers a cost-effective framework for practical reading comprehension applications, without the need of laborious distractor annotation and costly large-size models. 
\end{abstract}

\section{Introduction}

Reading comprehension assessment holds significant importance in the educational field. Typically, a reading comprehension 
sample consists of four components: passage, question, answer and multiple distractors. In recent years, while the cloze-style Distractor Generation (DG) task 
has received wide interest~\cite{DBLP:conf/aaai/RenZ21,DBLP:conf/emnlp/ChiangWF22,DBLP:conf/acl/WangHYTSHF23}, the DG task
with complete long sentences is less addressed, primarily due to a scarcity of available supervised data.

Limited by the expensive annotation cost, there are just a few reading comprehension datasets for 
DG
from examination scene~\cite{lai-etal-2017-race,DBLP:journals/tacl/SunYCYCC19}. Methods on these single-sourcing datasets~\cite{DBLP:conf/aaai/GaoBLKL19,DBLP:conf/aaai/ZhouLW20} all face challenges of insufficient generalization in real-world applications.
On the other hand, unsupervised generation methods remain inaccessible considering the difficulty brought by the reading comprehension context. 

\begin{figure}[t]
\centering
\includegraphics[width=\columnwidth]{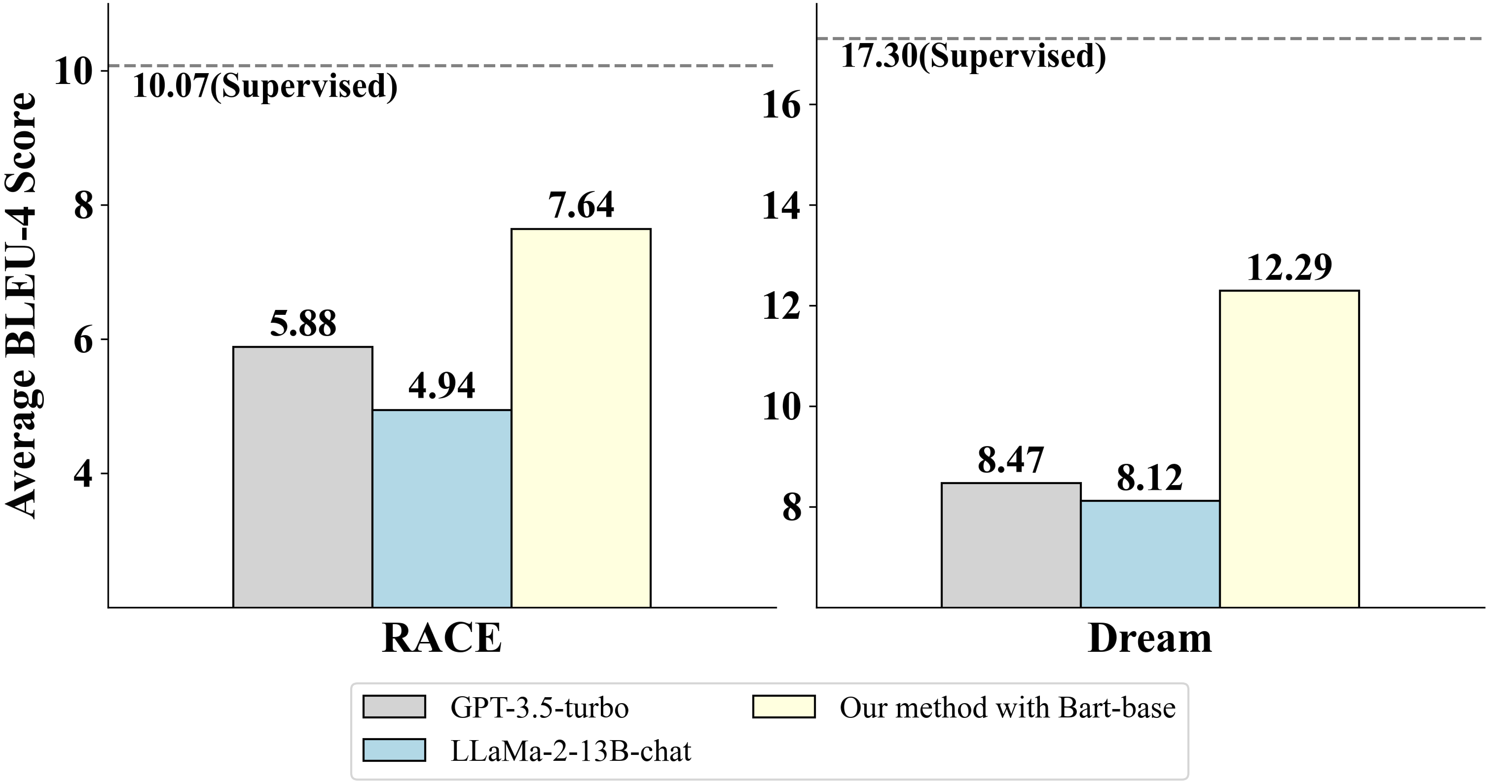}
\caption{Performance of different unsupervised methods 
generating 3 distractors on two datasets. We also display results with the supervised Bart-base model for comparison.} 
\label{fig:intro}
\end{figure}

Recently, LLMs like GPT~\cite{DBLP:journals/corr/abs-2303-08774} and LLaMa~\cite{touvron2023llama} have demonstrated powerful ability as automatic annotators to label 
training data~\cite{arora2022ask, gilardi2303chatgpt}. 
LLMs have been successfully 
applied in various fields of NLP, including multi-choice question-answering task~\cite{bitew2023distractor,nasution2023using,doughty2024comparative}. However, compared to previous 
fine-tuned methods, mainstream LLMs often fail to achieve a satisfactory performance on DG,
as 
illustrated in Figure~\ref{fig:intro}. Additionally, deploying LLMs in real world applications is challenging due to their substantial computational resource requirements and the closed-source model parameters.


To meet the high need of DG for real 
applications, where there are no distractor labelling data and only limited computational resource, we propose an unsupervised DG framework with a small model.
We 
adopt the distilling paradigm to enhance the smaller student model's generation capability with pseudo labels from LLMs~\cite{smith2022language,arora2022ask},
regarding LLMs as data annotators to assist the training of the smaller model~\cite{wang2021want}. 
Recognizing the suboptimal performance of LLMs on DG,
we 
propose a dual task training 
strategy by 
integrating both pseudo distractors and golden 
answers as training targets. 
Furthermore, we 
devise a two-stage training framework to reduce the negative impact caused by the conflict semantics presented in answers and distractors.
Note that we do not use the reference distractors for the unsupervised setting.

The utilization 
of answer information as training target, although improves the model's performance in generation quality, 
makes harm to the counterfactual capability of the generative model. To 
address this issue, we introduce 
contrastive decoding~\cite{DBLP:conf/acl/LiHFLEHZL23} into the inference process of 
DG. Specifically, we penalize factual text patterns favored by the answer generation module 
while encourage 
counterfactual results generated by the distractor generation agent.
Additionally, we 
apply plausibility constraint to restrict the effect of contrastive decoding for more stable generation results.
Note that we do not leverage LLMs during inference to ensure an easy deployment in real applications. 

We conduct experiments on RACE~\cite{lai-etal-2017-race} and Dream~\cite{DBLP:journals/tacl/SunYCYCC19}. As illustrated in Figure~\ref{fig:intro}, our unsupervised method with Bart-base significantly outperforms zero-shot LLMs 
with 200$\times$ fewer model parameters. 
Experimental results also show that our proposed counterfactual contrastive decoding method greatly improves the distracting capability of the generation model. Moreover, we leverage GPT-4~\cite{DBLP:journals/corr/abs-2303-08774} to evaluate the generated distractors, demonstrating that our method obtains a better performance than GPT-3.5-turbo both in generation quality and distracting level.    

To sum up, our contributions are as follows:

\begin{itemize}
    \item We propose an unsupervised 
    DG framework with dual task training, integrating the original answer information and pseudo distractors generated by LLMs. 
    \item We devise a new counterfactual 
    contrastive decoding method to improve the distracting level of generated outputs.  
    The proposed optimization can be transferred to other counterfactual generation tasks.
    \item Our method greatly outperforms teacher LLMs across various 
    evaluation metrics. 
    Our method provides a valuable 
    approach for constructing reading comprehension data in diverse real-world applications, 
    eliminating the need for costly human-annotated data and large-size models.
\end{itemize}

\section{Related Work}
\subsection{Distractor Generation}

Previous researches on distractor generation in reading comprehension mostly concentrate on designing attention framework based on end-to-end models~\cite{DBLP:conf/aaai/GaoBLKL19}. Co-attention~\cite{DBLP:conf/aaai/ZhouLW20} and reforming
modules~\cite{DBLP:conf/coling/QiuWF20} among passage, question and answer are proposed to extract key information about question-related and counterfactual details. Besides, Mixture of Expert is utilized in some works to ensure the quality and diversity of the generation results~\cite{DBLP:conf/coling/QiuWF20, qu2023accurate}.

Recently, 
DG studies with LLMs mostly focus on knowledge-based distractor generation in education field. \citet{bitew2023distractor} explores question-similarity based example selection method to enhance LLMs' DG 
performance in in-context learning. \citet{doughty2024comparative} designs complex prompt to generate multi-choice question answering data for Python programming learning. \citet{nasution2023using} asks ChatGPT to construct multi-choice data with the input of biology subject for biology learning.

\begin{figure*}[t]
\centering
\includegraphics[width=\linewidth]{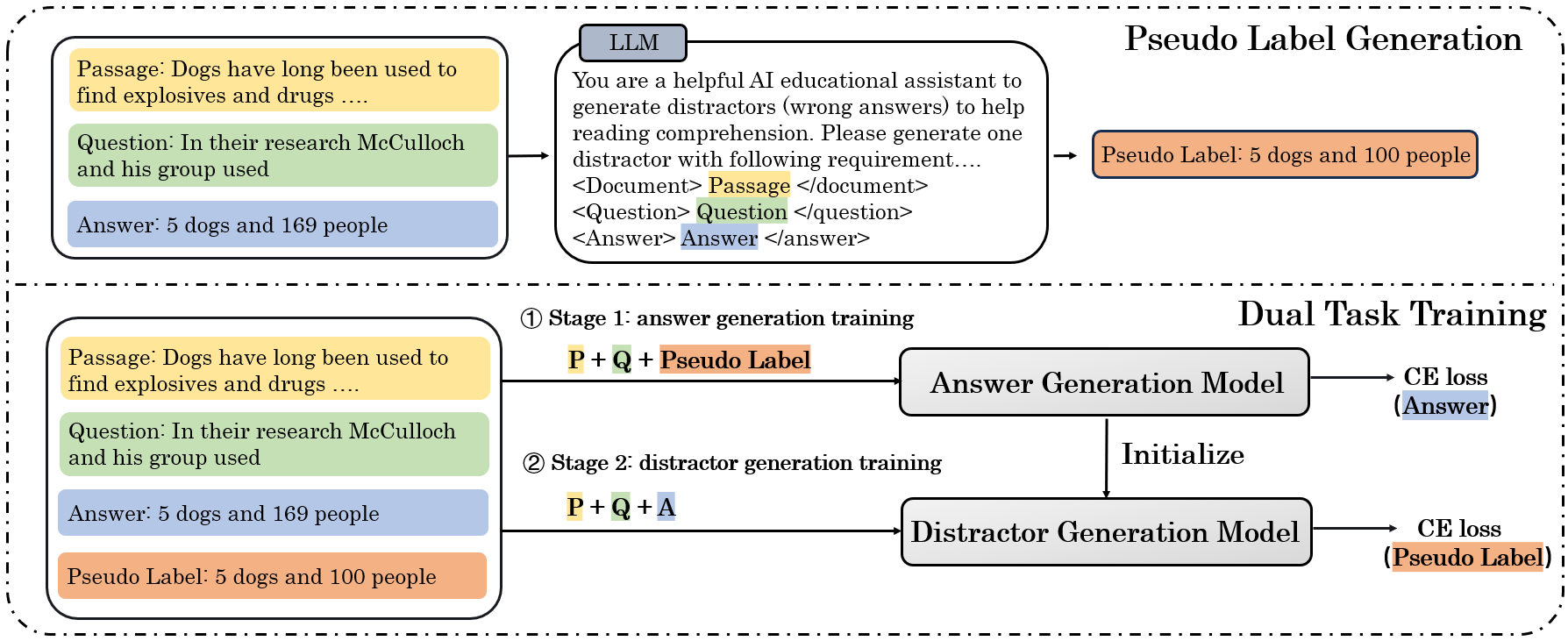}
\caption{Overview of our proposed unsupervised distractor generation 
framework, which 
can be divided into two parts: pseudo distractor generation and dual task training.} 
\label{fig:process}
\end{figure*}

\subsection{LLM Knowledge Distillation}
LLMs have been widely applied to generate pseudo labels to reduce the labeling cost in unsupervised situation. Recent works have proved the effectiveness of LLMs in annotating accuracy compared to human-annotated results in various NLP tasks~\cite{gilardi2303chatgpt,he2023annollm}. Related works have already been applied in question answering~\cite{saad2023udapdr}, information retrieval~\cite{bonifacio2022inpars}, text summarization~\cite{wang2021want} and common sense reasoning~\cite{whitehouse2023llm}
, covering both NLU and NLG fields.

Similar unsupervised methods are still under-explored in DG task, though previous works suffer from the expensive labeling cost. Addressing this gap is the main target of our paper.

\section{Method}

\subsection{Task Definition}

The reading comprehension data generally consists of four components: passage, question, answer and multiple distractors. 
In this context, the 
DG model regards the triplet of passage($p$), question($q$) and answer($a$) as input and generates results with a probability of $p_d$:

\begin{equation}
    p_d = p(d_i|p,q,a,d_{<i})
\end{equation}


In this work, we propose an unsupervised framework for DG task,
where the reference distractors in the training data are unseen but only the passage($p$), question($q$) and answer($a$) are provided. In this way, we would liberate experts from the laborious work of distractor annotation.

An overview of our unsupervised framework is displayed
in Figure~\ref{fig:process}. We first apply LLM as a teacher model to generation pseudo distractors. Next, we train a smaller student model with both pseudo labels and answers as 
generating targets through a two-stage training process.


\subsection{Generating Distractors with LLMs}
\label{sec:aug}
Previous works on DG task~\cite{DBLP:conf/aaai/GaoBLKL19,DBLP:conf/aaai/ZhouLW20} mostly depend on human-annotated data. The common-used dataset like RACE~\cite{lai-etal-2017-race} is sourced from the educational domain and annotated by professional teachers, exhibiting a high quality but expensive cost. The powerful generation capability of LLMs presents a chance for DG task in reading comprehension to overcome the problem of limited data.

Instead of utilizing human-annotated distractors, we obtain pseudo distractors from a LLM teacher with the input passage, question and answer. To save space, we display the prompts in Appendix~\ref{appendix:prompt_pseudo}. 

To guarantee the distracting level of the generated pseudo distractors, we filter out results that exhibit high similarity to the answer by calculating the BLEU-4 score. 
All pseudo 
outputs with a BLEU-4 score greater than 30 are dropped, along with their associated question-answer pairs.

\subsection{Dual Task Training with Student Models}

The pseudo distractors from LLMs (denoted as $d'$) can be directly 
distilled to a student model as supervised signals during training. However, employing this straightforward augmentation method may not lead to satisfactory 
results because of the suboptimal performance of LLM in DG task.

As a pair of dual tasks, the answer generation task exhibits similarities with the DG task. Both tasks require a comprehensive understanding on the input passage and deep analysis on question-related contexts. To this end,
we 
introduce answer generation into our unsupervised distractor
training process as an auxiliary task.

To take advantage of both pseudo distractor generation and answer generation, we devise a two-stage training procedure for the smaller student model. Firstly, the model treats [$p, q, d'$] as input and generates $a$, addressing 
answer generation: 

\begin{equation}
\label{eq:pa}
    p_a = p(a_i|p,q,d',a_{<i})
\end{equation}

Secondly, we just interchange answers and pseudo distractors, applying [$p, q, a$] as input and $d'$ as output, serving as 
distractor generation: 

\begin{equation}
\label{eq:pd}
    p_d = p(d'_{i}|p,q,a,d'_{<i})
\end{equation}


In the experiment, we prepend task-specific tokens to distinguish these two tasks. For distractor generation, we replace the decoder start token (often \texttt{eos} token) with the special token \texttt{[DIS]} and \texttt{[ANS]} for answer generation. 

We denote the model training in answer generation as $M_a$, and distractor generation as $M_d$. In the two-stage training procedure, $M_d$ is initialized with the parameters of $M_a$. Both models apply the cross-entropy (CE) loss for training.

\subsection{Contrastive Decoding}

The introduction of answer generation in dual task training raises up a challenge on the distracting level of the generated result, as the model may generate correct content. Inspired by the contrastive decoding method ~\cite{DBLP:conf/acl/LiHFLEHZL23}, we propose a novel decoding strategy called counterfactual contrastive decoding (CCD) to solve this issue.

\subsubsection{Counterfactual Contrastive Decoding} 

Generally speaking, CCD 
rewards counterfactual text patterns favoured by the distractor generation model 
while penalizes factual text patterns generated by the answer generation model. To obtain the output probabilities of both models, 
we propose a two-stage inference process:



\paragraph{Stage 1}
Perform 
the inference process with $p$, $q$, $a$ as input on 
DG model $M_d$ to generate one distractor result, 
referring as $d_{inter}$.

\paragraph{Stage 2}
Regard $M_d$ as the expert model and $M_a$ as the amateur model, and apply contrastive decoding as Equation~\ref{eq:ccd}:


\begin{equation}
\label{eq:ccd}
    \text{CCD-score}_i = \text{log} \frac{p_{d}(y_i|p, q, a, y_{<i})}{p_{a}(y_i|p, q, d_{inter}, y_{<i})}
\end{equation}
where $y$ is the output sequence of $M_d$. 

Please note that different from the training process,
$d_{inter}$ replaces the pseudo distractor $d'$ in the input sequence of $M_a$, 
thereby avoiding the dependence on LLM during inference. 



Taking into account the high similarity between $M_d$ and $M_a$ models, some common text patterns with high probability in both models will become hard to be generated 
while the generation probability of some implausible tokens will 
greatly improve.  
To address this issue, we optimize the counterfactual contrastive decoding as:





\begin{equation}
\begin{aligned}
\label{eq:ccd_scale}
    \text{CCD-score} &= \text{log-softmax}(\text{logit}'_d) \\
    \text{logit}'_d &= \text{logit}_d * f(\text{logit}_d, \text{logit}_a) \\
    \text{logit}_a &= M_a(p, q, d_{inter}, y_{<i}) \\
    \text{logit}_d &= M_d(p, q, a, y_{<i}) \\  
    \end{aligned}
\end{equation}

The scaling function $f$ is calculated as:
\begin{equation}
    f(x, y) = \text{exp}(\text{sgn}(x) * (\sigma(\frac{x - y}{t}) - 0.5)) 
\end{equation}
where $\sigma$ is the Sigmoid function, $\text{sgn}(*)$ is the Signum function. $t$ is a hyper-parameter to control the scaling degree and avoid the saturation of the sigmoid function.

As the final value of CCD-score mostly depends on the absolute value of $\text{logit}_d$, the implausible tokens with low generation probability are still hard to be generated after the logit scaling.
On the other hand, the scaling factor is determined by the difference between $\text{logit}_d$ and $\text{logit}_a$. Notably, it will be close to 1 when the $\text{logit}_d$ is approximately equal to $\text{logit}_a$, maintaining the high probability of tokens that both models exhibit high confidence on. 



\subsubsection{Plausibility Constraint}

To improve the stability of the decoding results, we introduce a plausibility constraint method to restrict the effect of counterfactual contrastive decoding.
For distractor generation, we propose a rank-based plausibility constraint:

\begin{equation}
    \text{CCD-score}=\left \{
    \begin{aligned}
    &\text{Equation~\ref{eq:ccd_scale}}  &\text{if} \ x_i \in \mathcal{V}_{adj} \\
    &\text{log-softmax}(\text{logit}_d)  &\text{otherwise} 
    \end{aligned}
    \right.
\end{equation}

\begin{equation}
\begin{aligned}
    &\mathcal{V}_{adj} = \{y_i \in \mathcal{V}: \ \text{rank}(p_d(y_i|y_{<i})) > r \}\\
\end{aligned}
\end{equation}

We rank the probability across the vocabulary. For tokens with extremely high probability, we will fix their logits and only adjust tokens ranked after $r$-th, where $r$ is a hyper-parameter.
Compared to the value-based plausibility constraint method proposed in ~\citet{DBLP:conf/acl/LiHFLEHZL23}, our rank-based method is more stable and controllable.

\section{Experimental Setup}

\subsection{Student Model Training}
We choose Bart-base as the student model
, which has only 139M parameters and can be easily deployed in real applications. 
We train the student model with 5-epoch answer generation and 10-epoch distractor generation in the two-stage training process. In both stages we set the maximum learning rate, batch size and warmup ratio to $10^{-5}$, 48 and 0.1. During inference, we adopt 
Jaccard Distance 
for generating 3 different results with the beam size as 20, and other detail settings can be referred to ~\citet{DBLP:conf/aaai/ZhouLW20}. $r$ and $t$ are set to 15 and 2 in experiment. 

\subsection{Applying LLMs}

We conduct experiments on two mainstream series LLMs: LLaMa-2 and GPT-3.5. For LLaMa-2, we apply LLaMa-2-13B-chat implemented by HuggingFace. For GPT-3.5, we apply GPT-3.5-turbo with the official API.

\paragraph{LLM distractor generation for distilling}
We apply the prompt shown in Appendix~\ref{appendix:prompt_pseudo} to generate one pseudo distractor for each input $p$, $q$ and $a$. For the sake of reproductivity, we turn off the sampling and set the temperature to zero. 

\paragraph{LLM zero-shot inference}

To explore the performance of LLM,
we apply LLMs to generate three 
distractors with the prompt displayed in Appendix~\ref{appendix:prompt_inf}. Due to the high cost of applying LLMs with a large beam size, we do not ask LLMs to return 20 results for filtering like student model does.

\subsection{Datasets}

We conduct extensive experiments on RACE~\cite{lai-etal-2017-race} and Dream~\cite{DBLP:journals/tacl/SunYCYCC19}, which are collected from the English exams of middle and high schools in China. The detailed statistics of the cleaned dataset are shown in Table~\ref{tab:data}.

\begin{table}[tp]
\renewcommand\arraystretch{1.1}
\setlength{\tabcolsep}{5pt}
\centering
\small
\begin{tabular}{c|c|c|c|c}
\hline
 & Dataset & \# q-a pair & \# passage & \# d(pseudo) \\
\hline
\multirow{2}*{\textbf{Train}} & RACE & 45120 & 20028 & 35452 \\
 ~ & Dream & 6116 & 3862 & 4721 \\
\hline
\multirow{2}*{\textbf{Test}} & RACE & 5787 & 2519 & - \\
~ & Dream & 2041 & 1283 &  - \\
\hline
\end{tabular}
\caption{The statistics of RACE and Dream dataset, where $q$, $a$ refers to the question and answer respectively. d(pseudo) is the pseudo distractor generated by GPT-3.5-turbo.}
\label{tab:data}
\end{table}

We drop the human-annotated distractors in training set and utilize these datasets as unsupervised data. LLMs are applied to generate one pseudo distractor for each question-answer pair in the training set for student model training. The generated results will be filtered with BLEU score as mentioned in Section~\ref{sec:aug}.

\subsection{Automatic Evaluation Metrics}

We apply BLEU\footnote{https://pypi.org/project/sacrebleu/}, Rouge\footnote{https://pypi.org/project/pyrouge/} and BertScore\footnote{https://github.com/huggingface/evaluate} to evaluate the generation quality, and Distinct~\cite{li2016diversity} to evaluate the generation diversity. 

As an important evaluation aspect, the distracting level of generated distractors has not received sufficient attention in previous works. Therefore, we propose a new automatic evaluation metric called \textbf{Faithful Score} to measure the distracting level of the generated results. 

\begin{table}[tp]
\renewcommand\arraystretch{1.1}
\centering
\small
\begin{tabular}{l|c|c}
\hline
 & \textbf{Dataset} & \textbf{Faithful Score}\\
\hline
\multirow{2}{*}{\textbf{Answer}} & RACE & 78.34 \\
~ & Dream & 76.23 \\
\hline
\multirow{2}{*}{\textbf{Distractor}} & RACE & 10.49 \\
~ & Dream & 10.81 \\
\hline
\end{tabular}
\caption{Faithful Score evaluation on the test set of RACE and Dream.}
\label{tab:fs}
\end{table}

\begin{table*}[tp]
\renewcommand\arraystretch{1.1}
\setlength{\tabcolsep}{4pt}
\centering
\small
\begin{tabular}{l|cccccc|cc|c}
\hline
 \textbf{Models} & \textbf{1-st B4} & \textbf{2-nd B4} & \textbf{3-rd B4} & \textbf{Avg B4} & \textbf{Avg BS} & \textbf{Avg R-L} & \textbf{Distinct 1} & \textbf{Distinct 2} & \textbf{Avg FS}($\downarrow$)\\
\hline
\multicolumn{10}{l}{\textbf{Fully Fine-tuned}} \\
\hline
HSA & 6.43 & 5.17 & 4.59 & 5.40 & - & 14.67 & - & - & -\\
HCA & 7.01 & 5.51 & 4.88 & 5.80 & - & 15.12 & - & - & -\\
EDGE & 7.57 & 6.27 & 5.70 & 6.51 & - & 18.27 & - & - & -\\
HMD-Net & 7.66 & 6.37 & 5.33 & 6.45 & - & 24.99 & - & - & -\\
MSG-Net & 8.87 & 8.86 & 8.53 & 8.75 & - & 26.39 & - & - & -\\
DG-MoE & 9.52 & 9.12 & 9.59 & 9.41 & 89.78 & 26.80 & 69.61 & 82.40 & - \\
\hline
Bart-base & 11.22 & 9.83 & 9.15 & 10.07 & 88.83 & 26.45 & 71.39 & 85.24 & 24.66 \\
\hline
\multicolumn{10}{l}{\textbf{Unsupervised}} \\
\hline
GPT-3.5-turbo & 6.82 & 5.75 & 5.07 & 5.88 & 86.73 & 26.13 & 75.46 & 87.68 & 33.39\\
LLaMa-2-13B-chat & 5.99 & 4.82 & 4.00 & 4.94 & 70.03 & 19.48 & 80.74 & 91.48 & 30.30 \\
Our Full Model & 8.36 & 7.43 & 7.14 & 7.64 & 88.38 & 26.38 & 70.33 & 83.86 & 25.64 \\
\hline
\end{tabular}
\caption{Experimental results on RACE dataset with 3 distractors comparing with baselines. B4 refers to BLEU-4, BS refers to BertScore, R-L refers to Rouge-L and FS refers to Faithful Score. The pseudo labels for model training are from GPT-3.5-turbo.}
\label{tab:main_result_race}
\end{table*}

\begin{table*}[tp]
\renewcommand\arraystretch{1.1}
\setlength{\tabcolsep}{4pt}
\centering
\small
\begin{tabular}{l|cccccc|cc|c}
\hline
 \textbf{Models} & \textbf{1-st B4} & \textbf{2-nd B4} & \textbf{3-rd B4} & \textbf{Avg B4} & \textbf{Avg BS} & \textbf{Avg R-L} & \textbf{Distinct 1} & \textbf{Distinct 2} & \textbf{Avg FS}($\downarrow$)\\
\hline
\multicolumn{10}{l}{\textbf{Fully Fine-tuned}} \\
\hline
Bart-base & 22.65 & 16.37 & 12.88 & 17.30 &  93.30 & 41.52 & 76.03 & 81.25 & 27.20 \\
\hline
\multicolumn{10}{l}{\textbf{Unsupervised}} \\
\hline
GPT-3.5-turbo & 10.91 & 8.18 & 6.33 & 8.47 & 91.80 & 33.00 & 71.73 & 77.26 & 28.67 \\
LLaMa-2-13B-chat & 11.81 & 7.63 & 5.92 & 8.12 & 87.13 & 31.01 & 80.61 & 82.91 & 24.22 \\
Our Full Model & 15.83 & 11.09 & 9.93 & 12.29 & 92.07 & 38.44 & 77.64 & 81.71 & 23.83 \\
\hline
\end{tabular}
\caption{Experimental results on DREAM dataset with 3 distractors. 
The pseudo labels are from GPT-3.5-turbo.}
\label{tab:main_result_dream}
\end{table*}

\paragraph{Faithful Score}
Based on RACE and Dream, we follow \citet{DBLP:journals/corr/abs-2011-03292} and train an Alberta model on the machine reading comprehension task. This model aims to judge whether a given candidate is a correct answer to the corresponding passage-question pair and return a classification score 
ranging [0, 100]. The 
DG task 
prefers models that
generate distracting results with low Faithful Score. 

We conduct evaluation on the test set of RACE and Dream, and results are shown in Table~\ref{tab:fs}. The Faithful Score values on reference answers are between 75 and 80 for both datasets and about 10 on distractors. The experiment results prove the excellent discrimination ability of this metric.







\begin{figure*}[t]
\centering
\subfigure[BLEU-4 Score]{ 
\label{fig:subfig_bleu} 
\includegraphics[width=0.85\columnwidth]{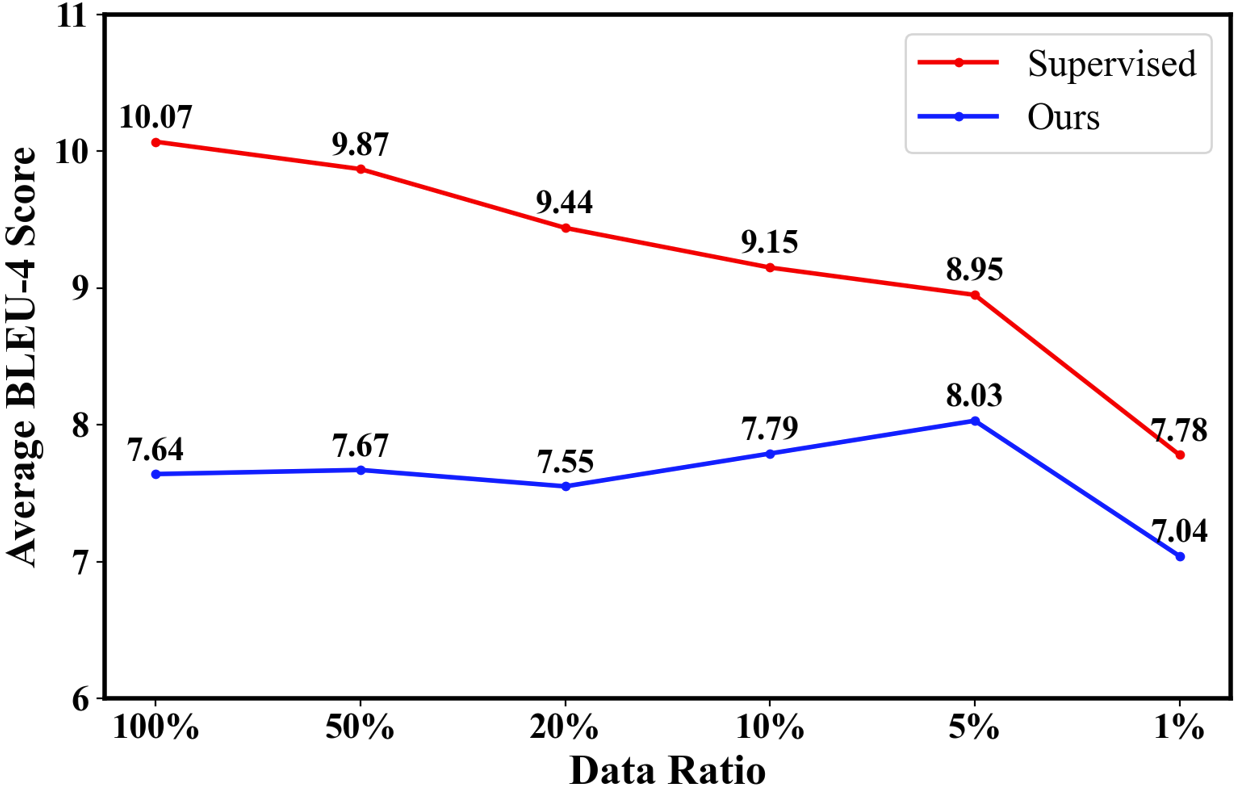}} 
\hspace{0in} 
\subfigure[Faithful Score]{ 
\label{fig:subfig_fs} 
\includegraphics[width=0.85\columnwidth]{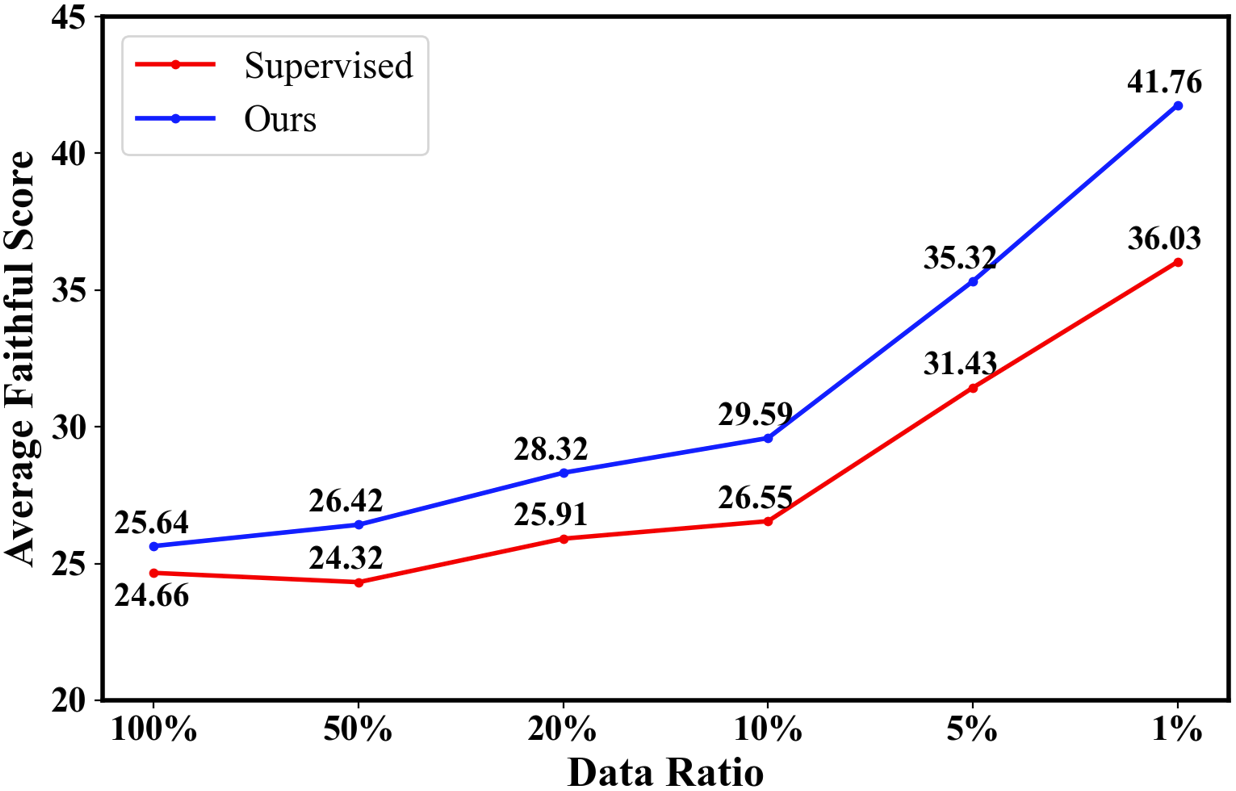}} 
\caption{Low-resource experimental results on RACE.} 
\label{fig:low_resource}
\end{figure*}

\section{Results and Analysis}

\subsection{Main Results}

The experiment results on RACE and Dream are shown in Table~\ref{tab:main_result_race} and Table~\ref{tab:main_result_dream}. For RACE, we display the results on previous works including HSA~\cite{DBLP:conf/aaai/GaoBLKL19}, HCA~\cite{DBLP:conf/aaai/ZhouLW20}, EDGE~\cite{DBLP:conf/coling/QiuWF20}, HMD-Net~\cite{DBLP:conf/cikm/MauryaD20}, MSG-Net~\cite{DBLP:journals/taslp/XiePCWH22}, DG-MoE~\cite{qu2023accurate}. The results of fine-tuned Bart-base and LLMs are also displayed for comparison. For our full model, we apply GPT-3.5-turbo to generate pseudo distractors. We mainly evaluate the results from three aspects: quality, diversity and distracting level.

\paragraph{Performance on the generation quality} 
Both GPT-3.5-turbo and LLaMa-2-13B-chat just manage to obtain half of the SOTA BLEU-4 score on two datasets. Compared to GPT-3.5-turbo, 
LLaMa-2-13B-chat is more unstable. On the more challenging dataset RACE, LLaMa-2-13B-chat achieves far inferior performance on BertScore (70.03) and Rouge-L (19.48) than other methods. 

Our proposed unsupervised method greatly outperforms LLMs on the generation quality. 
In terms of BLEU-4, we achieve 1.76 and 3.82 points improvements on RACE and Dream compared to GPT-3.5-turbo. And our model has achieved an approximate performance on BertScore and Rouge-L compared to the fully fine-tuned SOTA result. As for BLEU-4, there still exists an obvious gap between the unsupervised and fine-tuned results.

\begin{table*}[tp]
\renewcommand\arraystretch{1.1}
\centering
\small
\begin{tabular}{l|c|c|c|c|c|c}
\hline
 \textbf{Models} & \textbf{Avg B4} & \textbf{Avg BS} & \textbf{Avg R-L} & \textbf{Distinct 1} & \textbf{Distinct 2} & \textbf{Avg FS}($\downarrow$)\\
\hline
 Two-stage training & 7.75 & 88.48 & 26.77 & 69.34 & 83.37 & 27.59 \\
 \quad + Counterfactual Contrastive Decoding & 7.17 & 88.12 & 26.22 & 68.29 & 82.71 & 25.85 \\
 \quad \quad + Plausibility Constraint & 7.64 &  88.38 & 26.38 & 70.33 & 83.86 & 25.64 \\
\hline
\end{tabular}
\caption{Ablation studies on the contrastive decoding method.} 
\label{tab:cd_ablation}
\end{table*}

\begin{table}[tp]
\renewcommand\arraystretch{1.1}
\centering
\small
\begin{tabular}{l|c|c|c}
\hline
 \textbf{Models} & \textbf{Avg B4} & \textbf{Distinct 1} & \textbf{Avg FS}($\downarrow$)\\
\hline
Pseudo Label & 6.69 & 72.56 & 27.31 \\
Answer Label & 6.81 & 70.07 & 45.31 \\
Mixed Data 
& 8.02 & 69.16 & 32.68 \\
Two Stage & 7.75 & 69.34 & 27.59 \\
\hline
\end{tabular}
\caption{Experimental results on dual task training.}
\label{tab:dual_task}
\end{table}
\paragraph{Performance on the generation diversity}

We apply Distinct-1 and Distinct-2 to measure the generation diversity. On both RACE and Dream, our method achieves a close performance to SOTA. Among all these methods, LLaMa-2-13B-chat achieves greatest diversity performance. However, this may be due to the high randomness of its result that improves the generation diversity while makes harm to the generation quality~\cite{fang2022investigating}.

\paragraph{Performance on the distracting level}

We apply Faithful Score to evaluate the distracting level of the generated results. Our proposed method outperforms GPT-3.5 and LLaMa-2-13B-chat on both 
datasets and performs closely to the fully fine-tuned method. On RACE, our method is just 0.98 points lower than fine-tuned method, and on Dream our method outperforms 3.37 points.

\subsection{Effect of Dual Task Training}

We conduct experiments on RACE to investigate the impact of the dual task training based on GPT-3.5-turbo.
We train the student model in four different settings: with only pseudo distractors (Pseudo Label); with only answer targets (Answer Label); with the mixed data of pseudo distractors and answers (Mixed Data) and two-stage training process (Two Stage). 
In all these settings,
we do not apply our proposed counterfactual contrastive decoding 
during inference.

The results are shown in Table~
\ref{tab:dual_task}. Compared to pseudo-distractor-only method,
results of answer-only method have a slight improvement on BLEU-4 and a tiny decline on Distinct-1. However, the Faithful Score of pseudo-distractor-only method outperforms that of answer-only method significantly, with a decline of 18.0 points.
As for the dual task training with mixed data, the incorporation of answer information 
brings significant improvement on the generation quality. Nevertheless, it raises up a negative impact on the result's distracting level, as evidenced by 
the Faithful Score value.

The two-stage training process proves to be effective to address the above limitation.
We observe a decrease of 5.09 points in Faithful Score, accompanied by only a slight 0.27 points reduction in BLEU-4.

\subsection{Effect of Counterfactual Contrastive Decoding}

Further, we explore the effect of counterfactual contrastive decoding based on GPT-3.5-turbo. 
We apply the DG model with two-stage training as the base model. The results on RACE dataset are shown in Table~\ref{tab:cd_ablation}.

Despite a decline on the generation quality, CCD contributes to an improvement on model's counterfactual generation capability, with a 1.74 points decline on Faithful Score. Besides, plausibility constraint successfully enhances the stability of the generated results, further reducing Faithful Score by 0.21 points
and achieving a great trade-off between the generation quality and distracting level.



\subsection{Performance with Low-Resource Setting}

Our unsupervised method leverages answer information to enhance the generation quality. However, obtaining a sufficient number of annotated answer labels still requires investment, even if cheaper than distractors. In this section, we simulate the low-resource scenario of real-world applications on RACE dataset, and investigate the performance of both supervised and our unsupervised methods. The results 
are shown in Figure~\ref{fig:low_resource}. The experiments of our unsupervised method are conducted based on GPT-3.5-turbo.

As the data ratio of the training set decreases, the generation quality of the supervised method declines continuously. In contrast, our unsupervised method maintains a stable BLEU-4 score until the data ratio decreases to 1\% (about 400 samples) 
and demonstrates comparable performance to the supervised method in low-resource situation. 

With the decrease of the unannotated data number, the second stage training for distractor generation becomes insufficient, resulting in a heightened impact from the first stage answer generation training on the final results. 
This leads to a faster increase in the Faithful Score of unsupervised method compared to supervised one. 

Besides, we conduct experiments to explore LLMs' performance on few-shot settings. The results in Table~\ref{tab:fewshots} indicate that adding 5 randomly selected demonstrations to the prompt only increases the BLEU-4 score by 0.3 points for LLaMa-2-13B-chat.
\begin{table}[tp]
\renewcommand\arraystretch{1.1}
\centering
\small
\begin{tabular}{l|c|c|c}
\hline
 \textbf{Model} & \textbf{Shots} & \textbf{Avg B4} & \textbf{Avg FS}($\downarrow$)\\
\hline
\multirow{2}{*}{Llama-2-13b-chat}  & 0  & 4.94 & 30.30 \\
 & 5 & 5.24 & 28.76 \\
\hline
\end{tabular}
\caption{Experimental results on few-shot setting.}
\label{tab:fewshots}
\end{table}

\subsection{Evaluation from GPT-4}

N-gram based quality metrics like BLEU are not completely consistent with the actual performance in generation task. Thereby we apply GPT-4 as a professional evaluator to measure the performance of different methods from two aspects: the quality and distracting level of the generated results.

Concretely, we apply GPT-4 to compare two distractors generated by different methods. 
GPT-4 is asked to return "Win", "Lose" or "Tie" according to the comparison result. The order of two input 
distractors will be randomly shuffled in the input prompt for a fair comparison. We randomly select 1000 
samples from the test set of RACE for evaluation.
The prompt for GPT-4 evaluation is proposed in Section~\ref{appendix:prompt_evaluate}. The results are shown in Table~\ref{tab:evaluate}.

\begin{table}[tp]
\renewcommand\arraystretch{1.1}
\centering
\small
\begin{tabular}{l|ccc|ccc}
\hline
 \multirow{2}{*}{(\%)} & \multicolumn{3}{c|}{vs. Supervised} & \multicolumn{3}{c}{vs. GPT-3.5-turbo}  \\
 & Win & Tie & Lose & Win &Tie & Lose \\
\hline
Quality & 36.4 & 23.2 & 40.4 & 41.8 & 18.2 & 40.0 \\ 
Distracting & 47.1 & 12.8 & 40.1 & 49.1 & 7.5 & 43.4 \\ 
\hline
\end{tabular}
\caption{Evaluation results from GPT-4 with respect to the quality and distracting level of the generated distractors. We compare our unsupervised model with the supervised Bart-base and GPT-3.5-turbo zero-shot inference results.}
\label{tab:evaluate}
\end{table}

For the generation quality, GPT-4 favors results from the supervised method more, and our method's performance is slightly better than GPT-3.5-turbo. As for the distracting level,
our method performs significantly better than the other two methods, demonstrating the effectiveness of our counterfactual contrastive decoding strategy. 

\subsection{Performance of Different LLMs}

\begin{table}[tp]
\renewcommand\arraystretch{1.1}
\centering
\small
\begin{tabular}{l|c|c|c}
\hline
 \textbf{Teacher Models} & \textbf{Avg B4} & \textbf{Distinct 1} & \textbf{Avg FS}($\downarrow$)\\
\hline
GPT-3.5-turbo & 7.64 & 70.33 & 0.2564 \\
LLaMA-2-13B & 5.84 & 71.27 & 0.2583 \\
\hline
\end{tabular}
\caption{Experimental results with different large language models as the teacher models.}
\label{tab:llm_ablation}
\end{table}
Further, to explore the impact of different LLMs on the performance,
we 
adopt GPT-3.5-turbo and LLaMa-2-13B as different teacher models, and the results are shown in Table~\ref{tab:llm_ablation}.

Generally, the generation diversity and distracting level of the generated results are not significantly related to the selection of teacher model. However, the unsupervised method based on GPT-3.5-turbo outperforms LLaMa-2-13B by an average of 1.40 points in BLEU-4 score, which is positively correlated with the zero-shot performance of LLMs.

\begin{table*}[tp]
\centering
\small
\begin{tabularx}{\linewidth}{X}
\hline
\textbf{Passage}: Until late in the 20th century, most Americans spent time with people of generations. Now mid-aged Americans may not keep in touch with old people until they are old themselves .... The young, in turn, save the old. Once I was in a rest home when a visitor showed up with a baby ....  \\
\textbf{Question}: Now in an American family, people can find that \\
\textbf{Answer}: not all working people live with their parents \\
\hline
\textbf{Reference distractors}: \\
(1) children never live with their parents \quad (2) aged people are supported by their grandchildren \quad (3) grandchildren are supported by their grandparents
 \\
\textbf{Results of fully fine-tuned Bart-base}: \\
(1) all working people live with their parents \quad (2) parents don't care about their children \quad (3) all the old people live with their children \\
\textbf{Results of GPT-3.5-turbo}: \\
(1) all working people live with their parents \quad (2) working people live with their parents \quad (3) working people do not live with their parents \\
\textbf{Results of proposed unsupervised method}: \\
(1) only working people live with parents \quad (2) parents don't care about their children \quad
(3) the young save the old in turn \\
\hline
\hline
\textbf{Passage}: 'Where Are We Going , Dad ?' presents a new generation of men, in a break from Chinese tradition, now take an active role in their children's lives .... Another one must survive with his son for three days in the desert .... In traditional Chinese culture, the common conception of parenthood is that the father is strict and the mother is kind .... \\
\textbf{Question}: In Where Are We Going , Dad , we can probably learn about \\
\textbf{Answer}: how the fathers look after and guide their children \\
\hline
\textbf{Reference distractors}: \\
(1) how the fathers do housework at home \qquad (2) how the children study in their spare time \qquad (3) how the children help their father with their travel around China \\
\textbf{Results of fully fine-tuned Bart-base}: \\
(1) how the children grow up at the same time \quad (2) how the father is strict and the mother is kind \quad (3) how the fathers raise their children\\
\textbf{Results of GPT-3.5-turbo}: \\
(1) what kind of food the fathers and their children eat \quad (2) whether the fathers know how to do their daughters' hair \\
(3) how the fathers survive in the desert without cooking skills \\
\textbf{Results of proposed unsupervised method}: \\
(1) the negative impact of fathers on children's lives \quad (2) the history of Chinese culture \quad (3) why the father is strict \\

\hline
\end{tabularx}
\caption{Case study for fully fine-tuned model, GPT-3.5-turbo and our proposed unsupervised generation method.}
\label{tab:case}
\end{table*}

\subsection{Case Study}
\label{sec:case}
Table~\ref{tab:case} illustrates two examples of the generated distractors by three models. All three models produce fluent results without grammar errors. However, in case 1, the results from GPT-3.5-turbo suffers a problem of low diversity. And in case 2, both fine-tuned Bart-base and GPT-3.5-turbo generate \textbf{correct answers} like 'how the fathers raise their children', 'what kind of food the fathers and their children eat' and 'how the fathers survive in the desert without cooking skills'. This mistake is not observed in our unsupervised results with counterfactual contrastive decoding. 


\section{Conclusion}
In this paper, we propose an unsupervised distractor generation method. We apply Large Language Models as distractor labelers and construct dual task training process to enhance the student model's generation capability. Moreover, we optimize contrastive generation method in counterfactual generation context. Experiment results indicate that our method generally outperforms LLM-based approaches and is comparable to fully fine-tuned results in low-resource situations. Given the absence of human-annotated distractor dataset, our work can make contribution to building solid reading comprehension data in more future scenarios.  

\section*{Acknowledgements}
This work is supported by the National Natural Science Foundation of China (62076008) and the Key Project of Natural Science Foundation of China (61936012).

\section*{Limitations}

First, there still remains an obvious gap between our proposed unsupervised method and supervised method in 
DG task, especially for the quality of the generated results. Second, we only experiment on two English dataset from reading comprehension data due to the lack of high-quality testing data.
The performance of our method on other application scenarios requires further exploration. Third, we utilize human-annotated answer labels in our unsupervised method, which brings some cost of manual annotation. Fourth, while recent works succeed to enhance 
student models with rationales generated by LLMs in various NLU tasks, we fail to introduce these methods into 
DG task. We analyze that for NLG tasks, the rationales generated by LLMs rather complicate the generation process and put negative impact on the model's performance. Related work can be further explored in the future. Last, due to limited time, we do not explore the performance on more LLMs and student models with different parameter scales. 



\bibliography{custom}

\appendix

\section{Prompt for LLMs}
\label{appendix:prompt}
In this section we display the prompts applied to different LLMs.

\subsection{Prompt for generating pseudo distractors}
\label{appendix:prompt_pseudo}
We generate pseudo distractors by LLMs for the training of student model. The 
instructions are shown in Table~\ref{tab:prompt_pseudo}

\subsection{Prompt for zero-shot inference}
\label{appendix:prompt_inf}
When applying LLMs for zero-shot inference in distractor generation, we ask LLMs to generate three different results in one generation process. The prompt are shown in Table~\ref{tab:prompt_inf}

\subsection{Prompt for GPT-4 evaluation}
\label{appendix:prompt_evaluate}
We compare the quality and distracting level of results from two different models with GPT-4, and the prompts are shown in Table~\ref{tab:prompt_evaluate}.
\begin{table*}[tp]
\centering
\footnotesize
\begin{tabularx}{\textwidth}{X|X}
\hline
\multicolumn{1}{c|}{GPT-3.5-turbo} & \multicolumn{1}{c}{LLaMa-2-13B-chat} \\
\hline
"system": \newline
You are a helpful AI educational assistant to generate distractors (wrong answers) to help reading comprehension. Please generate one distractor with following requirement: 1. The generated distractor is a wrong answer to the input question according to the given document. 2. Return the generated result directly in one line that begin with '<result>' and end with '</result>'. \newline
"user": \newline
Now I will provide you with a reading comprehension document, a question and an answer. \newline
<document> {$p$} </document> \newline
<question> {$q$} </question> \newline
<answer> {$a$} </answer> & 
You are a helpful AI educational assistant to generate distractors (wrong answers) for reading comprehension. You are required to generate one distractor with the given document, question and answer. There are some requirements for you: 1. The generated result should begin with '<result>' and end with '</result>'. 2. If the input question is an incomplete sentence, the generated result should complete the syntax of the question. 3. You should not return any explanations except the distractors. There is the document, question and answer: \newline
<document> $p$ </document> \newline
<question> $q$ </question> \newline
<answer> $a$ </answer> \newline
<result>
\\
\hline
\end{tabularx}
\vspace{-1.0em}
\caption{Prompt for GPT-3.5-turbo and LLaMa-2-13B-chat to generate pseudo distractors.}
\label{tab:prompt_pseudo}
\end{table*}
\begin{table*}[tp]
\vspace{-0.3em}
\centering
\footnotesize
\begin{tabularx}{\textwidth}{X|X}
\hline
\multicolumn{1}{c|}{GPT-3.5-turbo} & \multicolumn{1}{c}{LLaMa-2-13B-chat} \\
\hline
"system": \newline
You are a helpful AI educational assistant to generate distractors (wrong answers) to help reading comprehension. Please generate three distractors with following requirement: 1. The generated distractors are a wrong answer to the input question according to the given document. 2. The generated results should be returned in three lines and each result should begin with '<result>' and end with '</result>'. \newline
"user": \newline
Now I will provide you with a reading comprehension document, a question and an answer. \newline
<document> {$p$} </document> \newline
<question> {$q$} </question> \newline
<answer> {$a$} </answer> & 
You are a helpful AI educational assistant to generate distractors (wrong answers) for reading comprehension. You are required to generate three distractors with the given document, question and answer. Now I will provide you with a document. \newline
<document> $p$ </document> \newline
There are some requirements for you: 1. The generated result should begin with '<result>' and end with '</result>'. Between <result> and </result>, return three results split by ';'. 2. If the input question is an incomplete sentence, the generated result should complete the syntax of the question. 3. You should not return any explanations except the distractors. Then I will give you a question-answer pair about the input document. \newline
<question> $q$ </question> \newline
<answer> $a$ </answer> \newline
The three distractors can be: <result>
\\
\hline
\end{tabularx}
\vspace{-1.0em}
\caption{Prompt for GPT-3.5-turbo and LLaMa-2-13B-chat to generate three different distractors.}
\label{tab:prompt_inf}
\end{table*}
\begin{table*}[tp]
\vspace{-0.3em}
\centering
\footnotesize
\begin{tabularx}{\textwidth}{X}
\hline
\multicolumn{1}{c}{Prompt for GPT-4} \\
\hline
"system": \newline
You are a helpful AI educational assistant that can evaluate distractors (wrong answers) and find the better one from two candidates. \newline
"user": \newline
Now I will provide you with a reading comprehension document, a question, an answer and a reference distractor. \newline
<document> {$p$} </document> \newline
<question> {$q$} </question> \newline
<answer> {$a$} </answer> \newline
<reference> {$d$} </reference> \newline
Then I will give you 2 distractor candidates and you should judge which one is a better result. The detailed comparison requirements are as follow: \newline
************ \newline
{$requirement$} \newline
************ \newline
I will show you two candidate distractors. If the first candidate is obviously greater than the second candidate, return 'Win'; If the first candidate is obviously worse than the second candidate, return 'Lose'; If you think there are not obvious gap between these two candidates, return 'Tie'. Do not return any explanations about your result. \newline
The candidates are: 1. {$candidates\_1$};  2. {$candidates\_2$}. \\
\hline
\end{tabularx} 

\begin{tabularx}{\textwidth}{X|X}
\hline
\multicolumn{1}{c|}{Requirements for quality evaluation} &  \multicolumn{1}{c}{Requirements for distracting level evaluation} \\
\hline
1. You should compare the candidates according to their quality. \newline
2. If the candidate is consist of fluent sentences without any grammar errors, the candidate has high quality. \newline
3. If there are just some small errors like tense error and voice error, the candidate has medium quality. \newline
4. If there are obvious syntactic or grammatical errors, the candidate has low quality. &
1. You should compare the candidates according to their distracting level. \newline
2.	If the candidate is correct to the input question, it has low distracting level. \newline
3.	If the candidate is wrong to the input question, it has high distracting level. \newline
4. The given answer has low distracting level and the given reference has high distracting level. These two sentences can serve as the reference for your comparison.
\\
\hline
\end{tabularx}
\vspace{-1.0em}
\caption{Prompt for GPT-4 to compare distractors generated by two different models.}
\label{tab:prompt_evaluate}
\end{table*}







\end{document}